\documentclass[preprint,12pt,authoryear]{elsarticle}

\usepackage{amssymb}
\usepackage{amsmath}

\usepackage{lineno}
\usepackage{rotating}

\usepackage{tikz}
\usetikzlibrary{positioning, arrows.meta, shapes.multipart}
\usetikzlibrary{shapes.geometric, arrows.meta, positioning}

\tikzstyle{startstop} = [rectangle, rounded corners,
    minimum width=3cm, minimum height=1cm, text centered,
    draw=black, thick, fill=gray!20]

\tikzstyle{process} = [rectangle,
    minimum width=3cm, minimum height=1cm, text centered,
    draw=black, thick, fill=blue!15]

\tikzstyle{decision} = [diamond, aspect=2,
    text centered, draw=black, thick, fill=orange!20]

\tikzstyle{arrow} = [thick,->,>=Stealth]

\usepackage{graphicx,booktabs,caption}
\usepackage{float} 
\usepackage{array}
\usepackage{multirow}
\usepackage[hidelinks]{hyperref}
\usepackage{subcaption}

\usepackage{gensymb}

\journal{Computer and Electronics in Agriculture}

\newcommand{\degC}{$^\circ\mathrm{C}$~}

\newcommand{\Ta}{$T_a$~}
\newcommand{\Tb}{$T_b$~}

\begin{document}

\begin{frontmatter}



\title{Multi-Step Forecasting of Grape Berry Temperature based on LSTM Model with Feed-Forward Attention}


\author{Srikanth Gorthi} 

\affiliation{organization={Center for Precision and Automated Agricultural Systems \& Department of Biological Systems Engineering, Washington State University},
            city={Prosser},
            state={WA},
            country={USA}}

\affiliation{organization={AgWeatherNet, Washington State University}, 
            city={Prosser},
            state={WA},
            country={USA}}

\author{L. G. Divyanth} 

\affiliation{organization={Center for Precision and Automated Agricultural Systems \& Department of Biological Systems Engineering, Washington State University},
            city={Prosser},
            state={WA},
            country={USA}}

\affiliation{organization={Department of Biological \& Environmental Engineering, Cornell University},
            city={Ithaca},
            state={NY},
            country={USA}}

\author{Dattatray Bhalekar} 

\affiliation{organization={Center for Precision and Automated Agricultural Systems \& Department of Biological Systems Engineering, Washington State University}, 
            city={Prosser},
            state={WA},
            country={USA}}

\author{Markus Keller} 

\affiliation{organization={Department of Viticulture and Enology, Washington State University},
            city={Prosser},
            state={WA},
            country={USA}}

\author{Lav Khot\corref{cor1}} 
\cortext[cor1]{Corresponding author.}
\ead{lav.khot@wsu.edu}

\affiliation{organization={Center for Precision and Automated Agricultural Systems \& Department of Biological Systems Engineering, Washington State University}, 
            city={Prosser},
            state={WA},
            country={USA}}

\affiliation{organization={AgWeatherNet, Washington State University},
            city={Prosser},
            state={WA},
            country={USA}}
\begin{abstract}
Accurate forecasting of grape berry temperature ($T_b$) is essential for enabling timely heat stress management in vineyards. In this study, a feed-forward attention mechanism integrated with a Long Short-Term Memory network (FAM-LSTM) was developed and evaluated for multi-step, high-resolution \Tb prediction. The models were trained using environmental data collected during the summer of 2023 and 2024 from a research vineyard (\textit{cv.} Chardonnay) located at Prosser, WA, USA, and validated on data from the summer of 2025. The proposed FAM-LSTM model was benchmarked against conventional approaches, including Long Short-Term Memory (LSTM), Gated Recurrent Unit (GRU), Recurrent Neural Network (RNN), and Random Forest (RF), across prediction horizons ranging from 15 minutes to 72 hours at a 15-minute temporal resolution (288 time steps). Additionally, two input scenarios were evaluated: (1) nearest open-field weather station observations and (2) in-vineyard microclimate measurements (i.e., air temperature and relative humidity). The model performance was assessed using mean absolute error (MAE), root mean squared error (RMSE), and mean absolute percentage error (MAPE). Results indicate that the FAM-LSTM model consistently outperformed all benchmark models across all prediction horizons and input scenarios. Furthermore, incorporating in-vineyard microclimate data significantly improved forecasting accuracy, particularly at longer horizons (72 h). Using open-field weather data, the FAM-LSTM achieved MAE, RMSE and MAPE in the respective ranges of 0.58 to 1.70\degree C, 0.65 to 2.07\degree C, and 2.65 to 8.09\%. In-vineyard observations further improved the model performance with respective MAE, RMSE, and MAPE in the ranges of 0.51 to 1.55\degree C, 0.71 to 1.87\degree C, and 2.48 to 7.68\%. Error analysis revealed that prediction uncertainty was highest during peak daytime periods (11:00 to 18:00) and increased progressively with the forecast horizon. Overall, FAM-LSTM framework offered a robust and high-resolution \Tb forecasting and could support precision heat stress management in vineyards. Future work will focus on extending the model to multiple grape cultivars and increasing forecast resolution to five minutes.    

\end{abstract}



\end{frontmatter}


\section{Introduction}

As in other parts of the world, heat waves (daily maximum temperatures above 35 \degree C) have become increasingly frequent in the Pacific Northwest (PNW) region of the United States. These high temperatures are affecting agricultural production, particularly in perennial specialty crops such as grapes, apples, and blueberries. Particularly in grapes, the optimal temperature to synthesize berry anthocyanin is approximately 30 \degC \citep{keller2010managing-56b}, and anthocyanin accumulation is highly sensitive to temperature and light during the ripening phase \citep{zheng2016use-abc}. The color, firmness, sugar, and acid content of grape berries are important indicators for their quality and they degrade when exposed to high air temperature and solar radiation for prolonged periods \citep{tarara2008berry-abc, cohen2012impact-abc, torres2020optimal-abc}. Thus, timely and effective heat-stress mitigation is crucial for maintaining grape berry quality. However, the lack of timely and reliable decision-support tools remain a significant challenge for implementing appropriate mitigation strategies.

Temperatures above 35 \degC with extensive solar radiation exposure can impose severe physiological stress on grape berries affecting growth, ripening, sugar accumulation, acidity, and coloration \citep{Keller0740033}. Prolonged exposure to heat and ultraviolet (UV) radiation can degrade anthocyanins in red cultivars and accelerate acid loss and dehydration in both red and white cultivars, ultimately diminishing wine quality. During a heat event, the berry temperature ($T_b$) can rise up to 15 \degC above ambient air temperature, with \Tb above 45 \degC potentially causing sunburn injury and associated yield and economic losses \citep{muller2023}. Therefore, accurate berry temperature estimation is critical for detecting heat stress on grapes for effective heat-stress mitigation in vineyards.

\par Berry temperature is governed by complex and nonlinear interactions among environmental variables, including air temperature ($T_a$), solar radiation, wind speed, relative humidity, and soil temperature ($T_s$) \citep{Heilman1994SoilAC, Hicks1973EddyFO, Nobel1999PhysicochemicalE, Kool2016EnergyAE}. For example, convective heat exchange between the surrounding air and berry surface can either increase or decrease \Tb depending on the prevailing temperature gradient and wind conditions, while incident solar radiation can substantially elevate \Tb above $T_a$. Previous studies have estimated \Tb using mechanistic or energy-balance models \citep{OrtegaFaras2010ParameterizationOA, Span2000EstimatingSA, Sene1994ParameterisationsFE, Cola2009BerryToneA, PoncedeLen2021A3M}, as well as physical berry mimics \citep{vogel2022prediction-3dc}. Although these approaches provide a physics-based representation of berry thermal dynamics, their performance is highly sensitive to boundary conditions and parameters such as aerodynamic resistance, surface emissivity, radiation exposure, and canopy architecture. These parameters vary with variable microclimates in vineyards and often lead to inaccurate berry temperature estimation.

\par Consequently, the application of physics-based models remains limited for real-time heat stress mitigation strategies in vineyards. Therefore, there is a need to develop a data-driven approach which can capture these non-linear environmental interactions while remaining adaptable to heterogeneous and dynamically changing vineyard conditions. In efforts to predict grape berry temperature, \citet{10948825} developed a machine-learning model using an in-field vineyard dataset; however, the model predicts the current berry temperature and does not have the ability to forecast future berry temperature. These berry temperature forecasts help in mitigation resource use planning and automated actuation of under-canopy cooling systems \citep{bhalekar2025grape, bhalekar2026berry}.  The latter systems have proven effective and model-driven actuation can bring precision to the actuation and help optimize irrigation water use in summer months. 
 
\par Advances in sensing technologies and Internet of Things (IoT)-based field monitoring now enable continuous environmental data collection at sub-canopy and cluster-level scales, facilitating the development of data-driven prediction models capable of learning nonlinear relationships among co-varying environmental variables \citep{Rajak2023InternetOT}. Traditional statistical and machine learning methods, such as linear regression, support vector regression (SVR), and autoregressive integrated moving average (ARIMA), have been applied to environmental forecasting but often struggle to capture temporal dependencies and perform poorly in long-term or multi-step prediction tasks \citep{Wang2022EfficientMT, Guo2020AMF, Tsai2020ApplicationOR, Dai2022ResearchOW, Kour2022ModellingAF}. For real-time heat stress mitigation in vineyards, mid- to long-term forecasting at high temporal resolution is critical. Specifically, forecasts covering 24 to 72 hours at 15-minute intervals would enable growers to schedule equipment deployment, procure water resources, and implement canopy interventions well before damaging temperatures are reached.

\par Recent progress in deep learning, such as convolutional neural networks (CNN), recurrent neural networks (RNN), gated recurrent units (GRU), and long short-term memory (LSTM) networks, have provided powerful tools for multivariate time-series modeling. For example, \cite{Codeluppi2021ForecastingAT} introduced a neural network framework for predicting greenhouse temperature using both historical and real-time temperature observations. \cite{Liu2022ACM} proposed the GCP-LSTM model, which leverages LSTM networks to capture nonlinear interactions among past environmental variables and provides short-term (5-min) greenhouse temperature forecasts. Although these studies illustrate the capability of deep learning to learn temporal patterns in controlled agricultural environments, they ignore environmental coupling effects, offer only short prediction windows, and do not sufficiently address multi-step forecasting. \cite{Moon2020PredictionOC} achieved reliable 30-minute greenhouse $CO_{2}$ prediction using an LSTM model, yet the performance dropped markedly for 2-hour forecasts.  Similarly, \cite{Choi2019PredictionOA} observed declining accuracy for extended multi-step predictions of greenhouse temperature and humidity when using a multilayer perceptron (MLP). 

\par The attention mechanism has recently shown remarkable success in both natural language processing and computer vision, owing to its strong feature extraction and long-range dependency modeling capabilities \citep{Mnih2014RecurrentMO}. In environmental prediction research, attention-based models are increasingly being adopted to enhance both accuracy and interpretability. \cite{Li2024ForecastingGA}, for example, designed an attention-augmented LSTM architecture capable of forecasting air and soil temperatures up to 480 minutes into the future with strong accuracy. Likewise, \cite{Raffel2015FeedForwardNW} proposed a streamlined feed-forward attention mechanism (FAM) that efficiently consolidates temporal features for multi-step time-series forecasting by computing raw attention scores directly from input representations, without requiring the computational overhead of query–key interactions. These developments demonstrate that FAM can substantially enhance both the precision and stability of sequential prediction models. Notably, while time-series deep learning has been applied to controlled environment agriculture and greenhouse conditions, research specifically targeting \Tb forecasting in open-field vineyard systems remains largely absent in the current literature. Moreover, standard weather data from open-field stations (e.g., regional or mesonet networks) are widely used for agricultural decision-making but may not fully represent site-specific microclimatic conditions, where berries experience dynamic radiative and convective thermal environments due to canopy architecture, foliage density, and row orientation. In contrast, in-vineyard weather measurements collected within or near the canopy can help capture the localized effects of canopy architecture and soil–plant–atmosphere interactions.  A comparative evaluation of models developed using open-field versus in-vineyard weather data is therefore essential to quantify the relative influence of data source on model accuracy and forecast reliability. Such an evaluation can provide valuable insights into whether readily available open-field data can serve as reliable surrogates for predicting $T_b$, or if localized monitoring is critical for reliable heat-stress forecasting.

\par This study was aimed towards developing accurate multi-step prediction of \Tb in vineyard \textit{(cv.} Chardonnay) systems to enable proactive heat-stress mitigation decision-making. In this research, the feed-forward attention mechanism was integrated with an LSTM model to assign learned, and dynamic attention weights to input features across different time steps, thereby improving the model's capacity to exploit the most informative historical data. The specific study objectives were to: 1) design and develop an FAM-LSTM architecture capable of forecasting \Tb at 15-minute intervals across a 72-hour horizon, and 2) conduct comparative experiments using open-field and in-vineyard weather data and benchmark FAM-LSTM against mainstream prediction models, including LSTM, RNN, GRU, and random forest (RF).

\section{Materials and Methods}

\subsection{Data Collection}
The data collection was performed in the Roza experimental vineyard of Washington State University (WSU) located in Prosser, WA, USA (46$^\circ$17$^\prime$32$^{\prime\prime}$N, 119$^\circ$44$^\prime$25$^{\prime\prime}$W). The site is classified as arid climatic type with hot, dry summers and cold winters. The vineyard (\textit{Vitis vinifera L. cv.} Chardonnay) was established in 2010 with 1.82 $\times$ 2.74 m vine to row spacing and north–south row orientation. The vines were trained to a bilateral cordon system, and shoots were vertically positioned between two pairs of foliage wires positioned at heights of 1.3 m and 1.7 m above the soil surface, separated by 0.3 m. Each year, pruning was done in late winter, and no further canopy management was practiced during the growing season. Drip irrigation was used to irrigate the vines following regional management practices of regulated deficit irrigation \citep{diverres2024response-abc}. Data were collected during the 2023, 2024, and 2025 growing seasons, specifically from 15 July to 30 August of each year, encompassing the critical period of rapid berry development, veraison (onset of ripening), and early ripening, when heat stress risk is greatest.

\begin{figure}[htbp]
\centering
\includegraphics[width=0.75\linewidth]{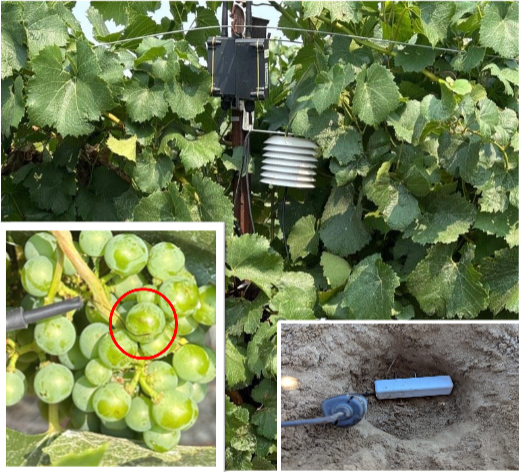}
\caption{The experimental setup to collect berry temperature (thermistor insertion, highlighted in red) and in-vineyard microclimate (air temperature and humidity and soil moisture probe) data using LoRaWAN sensing node.}
\label{fig:setup}
\end{figure}

Berry temperature was measured continuously using one thermistor (ST-200, Apogee Instruments, Logan, UT, USA) and three 'E-type' thermocouples inserted into the berries that were exposed to the afternoon sun on the west side of the canopy. Also, an air temperature and humidity probe (ATMOS 14, Meter Group Inc. Pullman, WA, USA) was installed at 1.5 m above ground level (AGL) and a soil moisture probe (TEROS 11, Meter Group Inc. Pullman, WA, USA) was installed at 10 cm depth as shown in Figure \ref{fig:setup}. The data were collected using a Long Range Wide Area Network (LoRaWAN) wireless sensing network \citep{gorthi2026lorawan} at 15 minute intervals. 

Two environmental datasets were prepared to evaluate the modeling framework under different sources of meteorological information. First, an in-vineyard weather (IW) dataset was compiled using measurements collected by sensors installed within the experimental vineyard block, including air temperature ($T_{a,\mathrm{IW}}$), relative humidity ($RH_{\mathrm{IW}}$), soil moisture ($SM_{\mathrm{IW}}$), and soil temperature ($T_{s,\mathrm{IW}}$). Second, an open-field weather (OW) dataset was obtained from the nearest WSU AgWeatherNet weather station and included air temperature ($T_{a,\mathrm{OW}}$), relative humidity ($RH_{\mathrm{OW}}$), solar radiation ($SR_{\mathrm{OW}}$), and wind speed ($WS_{\mathrm{OW}}$), measured at approximately 1.5 m AGL. The weather station was located $\simeq$500 m from the vineyard and at the same elevation. These two datasets were used to assess whether \Tb could be accurately predicted using readily available open-field weather observations compared with localized in-vineyard microclimate measurements. All sensor and weather-station observations were temporally synchronized to ensure consistent alignment between measured \Tb and the corresponding environmental predictors.

\subsection{Data pre-processing}
Missing and anomalous data arising from sensor malfunctions, transmission errors, or power interruptions can compromise the continuity of time-series measurements and degrade model performance. To address these issues, a structured preprocessing workflow was applied before model development. First, the distribution of each variable was examined using box-plot-based outlier detection. Values below or above the 25th and 75th percentiles by more than 1.5 times the interquartile range (IQR) were flagged as outliers \citep{hess273492023}. These values were removed from the dataset and subsequently treated as missing observations.

\par To ensure temporal consistency, a uniform 15-min time index was constructed to represent the expected sampling frequency over the entire monitoring period. All weather and environmental variables along with \Tb measurements were re-indexed to this timeline. Missing observations were reconstructed using time-based linear interpolation when fewer than three consecutive samples were absent. For longer gaps, values were imputed using the nearest observations recorded under comparable weather conditions, thereby avoiding unrealistic interpolation across extended time spans. Following gap filling, a Savitzky–Golay smoothing filter (window length = 4, polynomial order = 2) was applied to attenuate high-frequency sensor noise. This filtering strategy preserves the magnitude and timing of diurnal temperature extremes, which are essential for accurately characterizing berry heat-stress dynamics. In addition, four derived features were incorporated to enhance the model’s ability to capture thermal variations. To quantify transitions in ambient conditions, adjacent difference features were computed from the air temperature series. The first-order difference (Equation \ref{eq:1}) represents the instantaneous change in air temperature between consecutive time steps, while the second-order difference (Equation \ref{eq:2}) captures the change in the ambient temperature gradient, effectively describing the acceleration or deceleration of the heating process.

\begin{equation}
    \label{eq:1}
    \Delta T_t = T_t - T_{t-1}
\end{equation}

\begin{equation}
    \label{eq:2}
    \Delta^2 T_t = \Delta T_t - \Delta T_{t-1}
\end{equation}

To model periodic diurnal patterns, cyclical time features were also incorporated by encoding the hour of day using sine and cosine transformations (Equations \ref{eq:3} \& \ref{eq:4}). This sine and cosine encoding preserves the continuous and cyclical nature of time, avoiding the artificial discontinuities at daily boundaries (e.g., between 23:45 and 00:00) and enabling the model to learn heating–cooling cycles that strongly shape both environmental variables and \Tb dynamics.

\begin{equation}
    \label{eq:3}
    hr_{sin} = sin(\frac{2\pi. hour}{24})
\end{equation}

\begin{equation}
    \label{eq:4}
    hr_{cos} = cos(\frac{2\pi. hour}{24})
\end{equation}

To reduce scale-related biases among multivariate environmental features and to facilitate model convergence, all variables were normalized using min–max scaling to a range of [0, 1]. The normalization function is defined as shown in Equation \ref{eq:5}:

\begin{equation}
    \label{eq:5}
    x^\prime = \frac{x - x_{min}}{x_{max} - x_{min}}
\end{equation}

where, $x$ is the original value of a given variable, $x^\prime$ is the normalized value, and $x_{max}$ and $x_{min}$ denote the maximum and minimum values of the variable across the entire dataset, respectively.

\subsection{Selection of input variables}

To avoid computational overhead from high-dimensional inputs, a feature selection process was performed prior to model training. Multi-dimensional environmental features can substantially increase model complexity and computation time. Therefore, the linear dependence between \Tb and each environmental variable used in both the in-vineyard and open-field models was evaluated. The variables included in-vineyard air temperature ($T_{a,IW}$), relative humidity ($RH_{IW}$), soil temperature ($T_{s, IW}$), soil moisture ($SM_{IW}$), $\Delta T_t$, $\Delta^2T_t$, $hr_{sin}$, and $hr_{cos}$ for the in-vineyard dataset, and open-field air temperature ($T_{a,OW}$), relative humidity ($RH_{OW}$), solar radiation ($SR_{OW}$), wind speed ($WS_{OW}$) for the open-field dataset. 

\par The Pearson correlation coefficient (\textit{r}) was employed to quantify the correlation degree between \Tb and each environmental factor. Following the classification criteria proposed by \cite{Schober2018CorrelationCA}, the strength of correlation between \Tb and each environmental factor was categorized as, 
very weak ($0 < |\mathrm{\textit{r}}| \leq 0.2$), 
weak ($0.2 < |\mathrm{\textit{r}}| \leq 0.4$), 
moderate ($0.4 < |\mathrm{\textit{r}}| \leq 0.6$), 
strong ($0.6 < |\mathrm{\textit{r}}| \leq 0.8$), 
or very strong ($0.8 < |\mathrm{\textit{r}}| \leq 1$). Environmental variables exhibiting very weak correlations were excluded from further modeling.

\subsection{Data preparation for time-series modeling}

To enable multi-step forecasting of $T_b$, the pre-processed time-series data with the selected variables were transformed into a supervised learning structure using a sliding-window approach. Specifically, a fixed-length input window $n_{in}$ and output window $n_{out}$ were defined. For each time step $t$, the input sequence comprised historical measurements from time steps $t-(n_{in}+1)$ to $t$, while the corresponding output sequence contained the future values of the \Tb from $t+1$~ to $t+n_{out}$ as shown in Equation \ref{eq:6},

\begin{equation}
    \label{eq:6}
    X_t = [x_{t-(n_{in}+1)}, x_{t-(n_{in}+2)}, \ldots, x_t], \\
    y_t = [y_{t+1}, y_{t+2},\ldots, y_{t+n_{out}}]
\end{equation}

where, $X_t$ represents the historical feature values over the past $n_{in}$ time steps, and $y_t$ denotes the multi-step \Tb forecasting targets over the next $n_{out}$ time steps. The sliding window was moved across the entire time series in 15-minute increments (i.e., one step), generating overlapping input-output pairs for both the in-vineyard and open-field modeling scenarios. Data from the 2023 and 2024 seasons ($n=2232$) were used as training set (75\%), while the 2025 season ($n=744$) data served exclusively as the test set (25\%).

\subsection{Grape berry temperature forecasting models}

\subsubsection{Long short term memory network}

Long short-term memory networks (LSTM) are an advanced form of recurrent neural networks (RNNs) that employ a cell state and gating mechanisms to mitigate the vanishing gradient problem and effectively capture nonlinear temporal patterns and long-term dependencies, making them well suited for \Tb prediction \citep{Liu2022ACM}. An LSTM cell contains three gating mechanisms, namely, the input gate ($i_t$), forget gate ($f_t$), and output gate ($o_t$), which regulates the flow of information through the network. The input gate determines how much information from the current input ($x_t$) and previous hidden state ($h_{t-1}$) is incorporated into the cell memory. The forget gate controls the retention or removal of information from the previous cell state ($c_{t-1}$). The output gate regulates how much of the updated cell state contributes to the hidden state ($h_t$).

The cell state ($c_t$) serves as the internal memory of the LSTM, enabling the network to retain relevant information over extended time periods. This gating structure enables selective retention and removal of information, making LSTMs particularly suitable for modeling
nonlinear dependencies in environmental time-series data. The operations of an LSTM cell at time step $t$ are given by Equations \ref{eq:lstm_input} -- \ref{eq:lstm_hidden}:

\begin{equation}
i_t = \sigma\left(W_i x_t + U_i h_{t-1} + b_i\right)
\label{eq:lstm_input} 
\end{equation}

\begin{equation}
f_t = \sigma\left(W_f x_t + U_f h_{t-1} + b_f\right)
\label{eq:lstm_forget}
\end{equation}

\begin{equation}
o_t = \sigma\left(W_o x_t + U_o h_{t-1} + b_o\right)
\label{eq:lstm_output} 
\end{equation}

\begin{equation}
\tilde{c}_t = \tanh\left(W_c x_t + U_c h_{t-1} + b_c\right)
\label{eq:lstm_candidate}
\end{equation}

\begin{equation}
c_t = f_t \odot c_{t-1} + i_t \odot \tilde{c}_t
\label{eq:lstm_cell}
\end{equation}

\begin{equation}
h_t = o_t \odot \tanh(c_t)
\label{eq:lstm_hidden}
\end{equation}

where $i_t$, $f_t$, $o_t$, $c_t$, $\tilde{c}_t$, and $h_t$
represent the input gate, forget gate, output gate, cell state,
candidate cell state, and hidden state at time step $t$, respectively.
$W_i$, $W_f$, $W_o$, and $W_c$ denote the weight matrices associated
with the current input, while $U_i$, $U_f$, $U_o$, and $U_c$ denote
the weight matrices associated with the previous hidden state.
The corresponding bias vectors are denoted by $b_i$, $b_f$, $b_o$,
and $b_c$. The functions $\sigma(\cdot)$ and $\tanh(\cdot)$ represent
the sigmoid and hyperbolic tangent activation functions, respectively,
and $\odot$ denotes element-wise multiplication.

\subsubsection{Feed-forward attention mechanism}

Inspired by advances in neural attention models in natural language processing and time-series prediction tasks, the feed-forward attention mechanism (FAM) was introduced to enhance the model’s ability to focus on critical information within long temporal sequences. The attention mechanism selectively concentrates on the most relevant parts of the input, allowing the model to capture salient features and dependencies more effectively. It has demonstrated remarkable success in various domains, including speech recognition, language modeling, and environmental time-series forecasting
\citep{Liu2022ACM, qin2017dual-stage-d20}. Unlike recurrent attention approaches that rely on query-key
interactions, the FAM computes importance
scores directly from the input features, making it computationally
efficient and well suited for real-time environmental modeling. For a given time step $t$, consider
$n$ input feature elements, represented as
$\mathbf{x}_t = \{x_{t,1}, x_{t,2}, \ldots, x_{t,n}\}$. A fully
connected feed-forward network computes a raw attention score
$e_{t,i}$ for each input feature:

\begin{equation}
e_{t,i} = \mathbf{v}^{T}
\tanh\left(\mathbf{W}x_{t,i} + \mathbf{b}\right)
\label{eq:fam_score}
\end{equation}

where $\mathbf{W}$ and $\mathbf{b}$ denote the weight matrix and bias
vector of the attention layer, respectively, and $\mathbf{v}$ is a
learnable attention vector.

The raw attention scores are subsequently normalized using the softmax
function to obtain the attention-weight distribution $\alpha_{t,i}$
across all input features at time step $t$:

\begin{equation}
\alpha_{t,i}
=
\frac{\exp(e_{t,i})}
{\displaystyle\sum_{j=1}^{n}\exp(e_{t,j})},
\qquad i=1,2,\ldots,n
\label{eq:fam_attention}
\end{equation}

where $\alpha_{t,i}$ represents the normalized attention weight assigned
to the $i$th input feature at time step $t$.

Finally, the context vector $\mathbf{z}_t$ at time step $t$ is computed
as the attention-weighted representation of the input features:

\begin{equation}
\mathbf{z}_t
=
\sum_{i=1}^{n}
\alpha_{t,i} x_{t,i}
\label{eq:fam_context}
\end{equation}

This formulation enables the FAM to dynamically adjust the contribution
of individual input features by assigning greater weights to the most
informative elements while down-weighting features that contribute less to the prediction.

\subsubsection{FAM-LSTM model configuration}

This study implemented a time series forecasting model for \Tb by combining LSTM and FAM. Initially, multivariate time series data features were memorized using the LSTM model. Subsequently, FAM was used to describe the relationship between future time steps and historical data at each time step. Several attention weights were assigned to indicate the degree of correlation between the current and historical time steps. Then, the major effects were extracted by allocating various weights to the input at each time step and weighted to sum.
The model receives multivariate time-series inputs formatted as a three-dimensional tensor $[S,T,X]$, where $S$ is the number of samples, $T=T_1, T_2, \ldots, T_i$ denotes the time-step window, and $X=X_1, X_2, \ldots, X_j$ represents the feature dimensions. These inputs are processed by two stacked LSTM layers, each followed by a dropout layer to reduce overfitting, allowing the network to encode long-range temporal dependencies and generate a sequence of hidden representations $H = h_1, h_2, \ldots, h_m$. The FAM computes a set of adaptive weights $\alpha = \alpha_1, \alpha_2, \ldots, \alpha_m$, where each $\alpha_m$ reflects the importance of the corresponding hidden state $h_m$. The dynamic weight $\alpha$ is applied to integrate the temporal information of the input sequence, forcing the model to focus on the characteristics of the important time steps. The weighted sum of hidden states forms the context vector, which captures the dominant temporal characteristics influencing $T_b$. Further, the fully connected layer integrates the context vector and performs dimensionality transformation to map the aggregated temporal representation into the prediction space. The output layer produces \Tb forecasts in the tensor form $[O, H]$, where $O$ is the number of samples and $H= H_1, H_2, \ldots, H_k$ represents the forecast horizon (with $H_k$ denoting the predicted \Tb at the $k^{th}$ step ahead).

\subsubsection{Model evaluation and comparison}

The proposed FAM-LSTM model was benchmarked against commonly used machine learning and deep learning models, including LSTM, recurrent neural network (RNN), gated recurrent unit (GRU), and random forest (RF). All models were implemented in Python 3.7 using the TensorFlow deep learning framework. The architecture of each model was optimized through repeated experimental evaluation by varying the number of recurrent and fully connected layers, and the number of units in each layer. Specifically, for the LSTM-based models, the optimal architecture consisted of two LSTM layers with 128 and 64 units, respectively, followed by a fully connected layer with 96 neurons. The optimal GRU architecture comprised two GRU layers with 96 and 64 units, followed by a fully connected layer with 64 neurons. Similarly, the RNN architecture consisted of two recurrent layers with 128 and 64 units and a fully connected layer with 64 neurons. For all deep learning models, the Adam optimizer was employed with an initial learning rate of 0.001, and mean squared error (MSE) was used as the loss function. A dropout rate of 0.2 was applied after each recurrent layer to reduce overfitting. Model training was performed for a maximum of 200 epochs, with early stopping based on validation loss using a patience of 20 epochs. For the RF model, the optimal configuration consisted of 300 decision trees with a maximum tree depth of 20 and MSE as the node-splitting criterion.

A 72-h input window, corresponding to 288 time steps at a 15-min temporal resolution, was used for all models. This window was selected to capture multiple diurnal cycles and associated temporal variations in environmental conditions while limiting unnecessary model complexity. Accordingly, all models used the preceding 72 h (288 time steps) of observations to generate multi-step-ahead \Tb predictions over a 72-h forecast horizon.

The performance of the prediction models was evaluated using three standard error metrics: mean absolute error (MAE), root mean square error (RMSE), and mean absolute percentage error (MAPE). MAE quantifies the average magnitude of prediction errors, providing a direct and interpretable measure of the overall deviation between predicted and observed \Tb values; lower MAE values indicate closer agreement with the observed values. RMSE applies a quadratic penalty to prediction errors, making it more sensitive to large deviations and therefore, useful for assessing the model's ability to avoid substantial prediction errors. MAPE expresses the prediction error as a percentage of the observed value, enabling scale-independent comparison across different forecasting horizons and environmental conditions. For all three metrics, lower values indicate higher predictive accuracy. The metrics are defined as follows (Equations \ref{eq:mae}-- \ref{eq:interval_error}):

\begin{equation}
\mathrm{MAE}
=
\frac{1}{N}
\sum_{i=1}^{N}
\left| y_i - \hat{y}_i \right|,
\label{eq:mae}
\end{equation}

\begin{equation}
\mathrm{RMSE}
=
\sqrt{
\frac{1}{N}
\sum_{i=1}^{N}
\left( y_i - \hat{y}_i \right)^2
},
\label{eq:rmse}
\end{equation}

\begin{equation}
\mathrm{MAPE}
=
\frac{100}{N}
\sum_{i=1}^{N}
\left|
\frac{y_i-\hat{y}_i}{y_i}
\right|,
\label{eq:mape}
\end{equation}

where $y_i$ and $\hat{y}_i$ represent the observed and predicted \Tb
values for the $i$th sample, respectively, and $N$ denotes the total
number of samples.

To examine model performance across multiple forecasting horizons, the
interval temperature prediction error ($E_{i,t}$) was used to quantify
the deviation between the predicted and observed \Tb at each future
time step $t$:

\begin{equation}
E_{i,t}
=
\hat{y}_{i,t} - y_{i,t},
\qquad
t = 1,2,\ldots,H,
\label{eq:interval_error}
\end{equation}

where, $\hat{y}_{i,t}$ and $y_{i,t}$ represent the predicted and observed
\Tb, respectively, for sample $i$ at forecasting time step $t$, and
$H$ denotes the total number of forecast horizons.

\section{Results and Discussion}

\subsection{Correlation analysis}

\begin{figure}[htbp]
    \centering
    \includegraphics[width=0.85\linewidth]{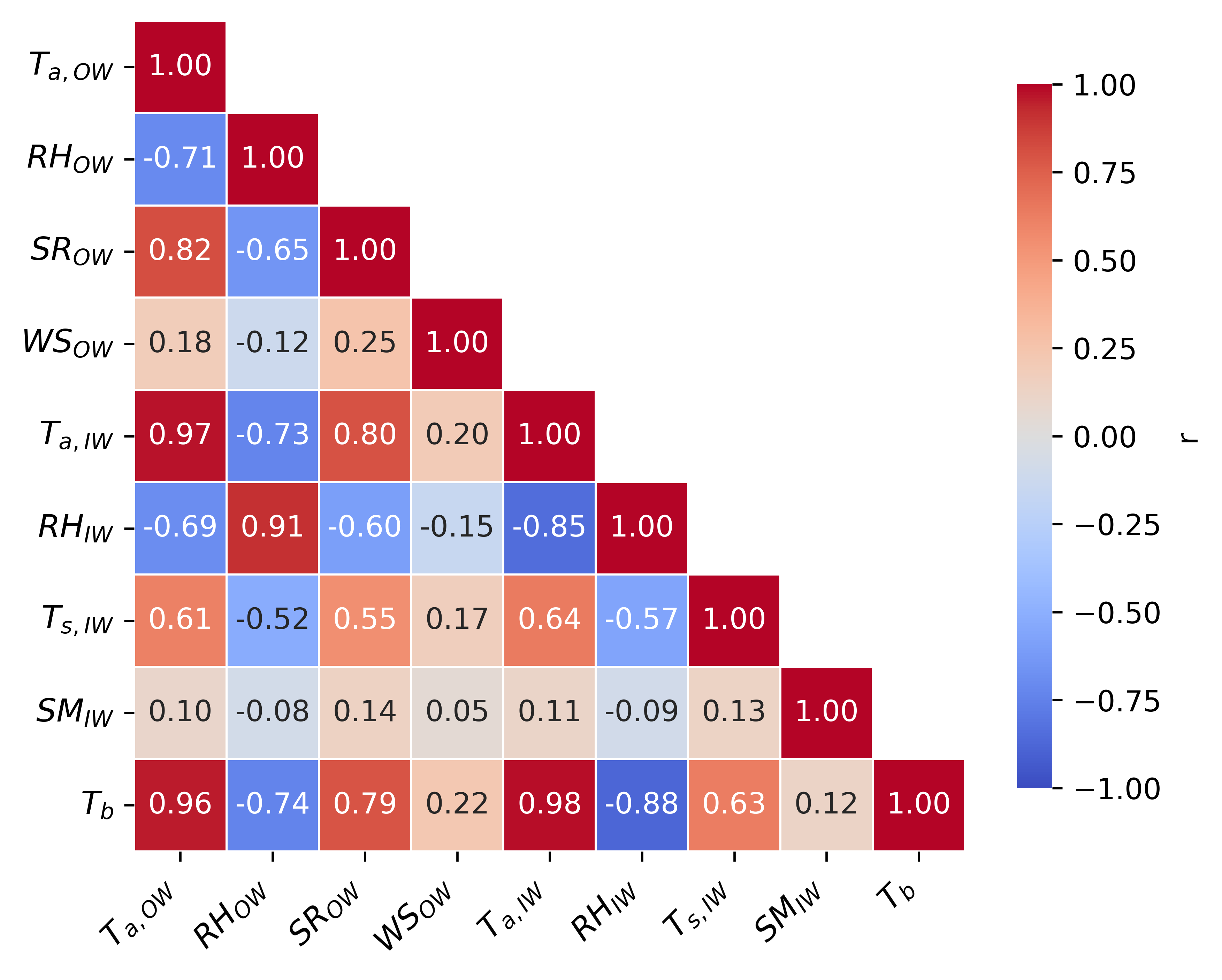}
    \caption{Pearson correlation analysis between the in-vineyard (IW) and open-field (OW) environmental datasets, including air temperature ($T_{a,\mathrm{IW}}$ and $T_{a,\mathrm{OW}}$), relative humidity ($RH_{\mathrm{IW}}$ and $RH_{\mathrm{OW}}$), soil temperature ($T_{s,\mathrm{IW}}$), soil moisture ($SM_{\mathrm{IW}}$), solar radiation ($SR_{\mathrm{OW}}$), wind speed ($WS_{\mathrm{OW}}$), and berry temperature ($T_b$). }
    \label{fig:corr}
\end{figure}

The Pearson correlation analysis between \Tb and environmental variables is depicted in Figure \ref{fig:corr} as a heatmap encompassing both open-field  and in-vineyard predictors. Among the open-field variables, air temperature ($T_{a,\mathrm{OW}}$; \textit{r} = 0.96) exhibited the strongest positive correlation with \Tb, followed by solar radiation ($SR_{\mathrm{OW}}$; \textit{r} = 0.79). These strong positive relationships indicate that ambient thermal conditions and incoming shortwave radiation are major drivers of berry surface warming \citep{smart1976solar-abc}. Relative humidity ($RH_{\mathrm{OW}}$; \textit{r} = $-0.74$) showed a strong negative correlation with \Tb, consistent with the inverse relationship between air temperature and relative humidity and their combined influence on the vineyard thermal environment and atmospheric vapor pressure deficit (VPD). In contrast, wind speed ($WS_{\mathrm{OW}}$; \textit{r} = 0.22) exhibited only a weak positive correlation with \Tb, suggesting a comparatively limited direct linear relationship with berry temperature under the observed conditions. Based on the predefined $|\mathrm{\textit{r}}| > 0.20$ selection threshold, all four open-field variables were retained as model inputs.

Among the in-vineyard variables, air temperature ($T_{a,\mathrm{IW}}$; \textit{r} = 0.98) exhibited the strongest correlation with \Tb, slightly exceeding that of open-field \Ta. This stronger association highlights the value of localized microclimate measurements, as in-vineyard air temperature more directly represents the thermal conditions surrounding the berries than measurements obtained from a standard open-field weather station. Similarly, in-vineyard relative humidity ($RH_{\mathrm{IW}}$; \textit{r} = $-0.88$) exhibited a stronger negative correlation with \Tb than its open-field counterpart, indicating that localized atmospheric moisture conditions were closely associated with berry thermal dynamics. As VPD drives berry transpiration, lower RH may be associated with greater evaporative cooling of the berries, even though this effect is likely limited due to the occlusion of stomata during berry development \citep{zhang2015grape-abc}. Soil temperature ($T_{s,\mathrm{IW}}$; \textit{r} = 0.63) showed a strong positive correlation with \Tb, potentially reflecting the coupling between soil and near-canopy thermal conditions throughout the diurnal cycle. In contrast, soil moisture ($SM_{\mathrm{IW}}$; \textit{r} = 0.12) exhibited a negligible linear correlation with \Tb and was therefore excluded from the in-vineyard model inputs. Soil moisture can indirectly influence canopy and berry temperatures through soil evaporation and plant transpiration, such effects may not be captured adequately by an instantaneous linear correlation.

In addition to the environmental predictors selected through correlation analysis, cyclical time features ($hr_{\sin}$ and $hr_{\cos}$) and the first- and second-order air-temperature differences ($\Delta T_t$ and $\Delta^2 T_t$) were retained as derived features in both modeling scenarios to represent diurnal periodicity and short-term temperature dynamics, respectively. Overall, the correlation analysis demonstrated stronger associations between \Tb and localized in-vineyard microclimate variables than their corresponding open-field measurements and provided the basis for selecting a parsimonious set of environmental predictors for subsequent \Tb modeling.

\subsection{Performance of open-field weather inputs in \Tb forecasting}

The prediction performance of the five models (RF, RNN, GRU, LSTM, and FAM-LSTM) using open-field weather inputs across seven forecast horizons is summarized in Table~\ref{tab:open_field_performance}. At the shortest forecast horizon ($t+1$, 15 min), FAM-LSTM achieved the lowest MAE (0.58~\degC), RMSE (0.65~\degC), and MAPE (2.65\%) among all evaluated models, indicating superior short-term \Tb prediction. The standard LSTM was the second-best model at this horizon (MAE = 0.65~\degC; MAPE = 3.16\%), followed by GRU (MAE = 0.68~\degC; MAPE = 3.46\%), RNN (MAE = 0.94~\degC; MAPE = 4.12\%), and RF (MAE = 1.05~\degC; MAPE = 5.35\%). The comparatively higher errors of RF indicate the advantage of recurrent architectures for this time-series prediction task, as their hidden states explicitly represent temporal dependencies within sequential environmental observations.

As the forecast horizon increased to 24 h ($t+96$), FAM-LSTM maintained its performance advantage, achieving an MAE of 1.47~\degC and MAPE of 7.12\%, compared with LSTM (MAE = 1.51~\degC; MAPE = 7.33\%), GRU (MAE = 1.69~\degC; MAPE = 8.19\%), RNN (MAE = 1.75~\degC; MAPE = 7.99\%), and RF (MAE = 2.36~\degC; MAPE = 11.99\%). The relatively moderate increase in prediction error suggests that incorporating the feature attention mechanism improved the model's ability to retain useful information from the historical input sequence as the prediction horizon increased. At the longest evaluated horizon of 72 h ($t+288$), FAM-LSTM continued to outperform the benchmark models, with an MAE of 1.70~\degC and MAPE of 8.09\%, compared with 1.84~\degC and 8.84\% for LSTM, 2.03~\degC and 9.64\% for GRU, 1.95~\degC and 9.36\% for RNN, and 2.45~\degC and 12.12\% for RF, respectively.

Overall, prediction accuracy decreased with increasing forecast time horizon across all models, reflecting the increasing difficulty of reproducing \Tb dynamics farther into the prediction window. Nevertheless, FAM-LSTM consistently achieved the lowest prediction errors across the reported horizons, demonstrating the benefit of augmenting the LSTM architecture with a feature attention mechanism. For FAM-LSTM, MAPE increased from 2.65\% at 15 min to 8.09\% at 72 h, corresponding to an approximately 205\% increase relative to the initial error. In comparison, LSTM MAPE increased from 3.16\% to 8.84\%, corresponding to an approximately 180\% increase. Thus, rather than indicating a smaller relative increase in error, the results show that FAM-LSTM maintained consistently lower absolute errors throughout the forecast horizon, including a 0.74 percentage-point reduction in MAPE relative to LSTM at 72 h.

\begin{table}[htbp]
\centering
\caption{Prediction performance of evaluated models using open-field weather inputs across multiple forecast time horizons.}
\label{tab:open_field_performance}

\setlength{\tabcolsep}{6pt}
\renewcommand{\arraystretch}{1.15}

\begin{tabular}{llccccccc}
\toprule
\textbf{Model} & \textbf{Metric}
& \textbf{$t+1$}
& \textbf{$t+4$}
& \textbf{$t+24$}
& \textbf{$t+48$}
& \textbf{$t+96$}
& \textbf{$t+192$}
& \textbf{$t+288$} \\

& 
& \textbf{15 min}
& \textbf{1 h}
& \textbf{6 h}
& \textbf{12 h}
& \textbf{24 h}
& \textbf{48 h}
& \textbf{72 h} \\
\midrule

\multirow{3}{*}{RF}
& MAE  & 1.05 & 1.59 & 1.76 & 1.86 & 2.36 & 2.43 & 2.45 \\
& RMSE & 1.44 & 2.05 & 2.25 & 2.41 & 3.01 & 3.11 & 3.14 \\
& MAPE & 5.35 & 7.78 & 8.43 & 9.21 & 11.99 & 12.04 & 12.12 \\
\addlinespace

\multirow{3}{*}{RNN}
& MAE  & 0.94 & 1.07 & 1.58 & 1.71 & 1.75 & 1.86 & 1.95 \\
& RMSE & 1.14 & 1.36 & 1.98 & 2.07 & 2.14 & 2.26 & 2.36 \\
& MAPE & 4.12 & 4.95 & 7.58 & 7.95 & 7.99 & 8.85 & 9.36 \\
\addlinespace

\multirow{3}{*}{GRU}
& MAE  & 0.68 & 1.01 & 1.53 & 1.72 & 1.69 & 1.94 & 2.03 \\
& RMSE & 0.81 & 1.29 & 1.88 & 2.11 & 2.06 & 2.31 & 2.38 \\
& MAPE & 3.46 & 4.99 & 7.54 & 8.36 & 8.19 & 9.23 & 9.64 \\
\addlinespace

\multirow{3}{*}{LSTM}
& MAE  & 0.65 & 0.94 & 1.46 & 1.62 & 1.51 & 1.75 & 1.84 \\
& RMSE & 0.74 & 1.13 & 1.79 & 1.97 & 1.88 & 2.20 & 2.20 \\
& MAPE & 3.16 & 4.53 & 7.29 & 7.94 & 7.33 & 8.40 & 8.84 \\
\addlinespace

\multirow{3}{*}{\textbf{FAM-LSTM}}
& MAE  
& \textbf{0.58}
& \textbf{0.86}
& \textbf{1.42}
& \textbf{1.57}
& \textbf{1.47}
& \textbf{1.69}
& \textbf{1.70} \\

& RMSE 
& \textbf{0.65}
& \textbf{0.98}
& \textbf{1.71}
& \textbf{1.85}
& \textbf{1.74}
& \textbf{2.06}
& \textbf{2.07} \\

& MAPE 
& \textbf{2.65}
& \textbf{4.20}
& \textbf{7.04}
& \textbf{7.79}
& \textbf{7.12}
& \textbf{7.91}
& \textbf{8.09} \\

\bottomrule
\end{tabular}

\vspace{2mm}

\footnotesize
\textit{Note:} MAE and RMSE are expressed in \degC; MAPE is expressed in \%.
Bold values indicate the best performance for each metric and forecast horizon.

\end{table}

\begin{figure}[htbp]
    \centering
    \includegraphics[width=\linewidth]{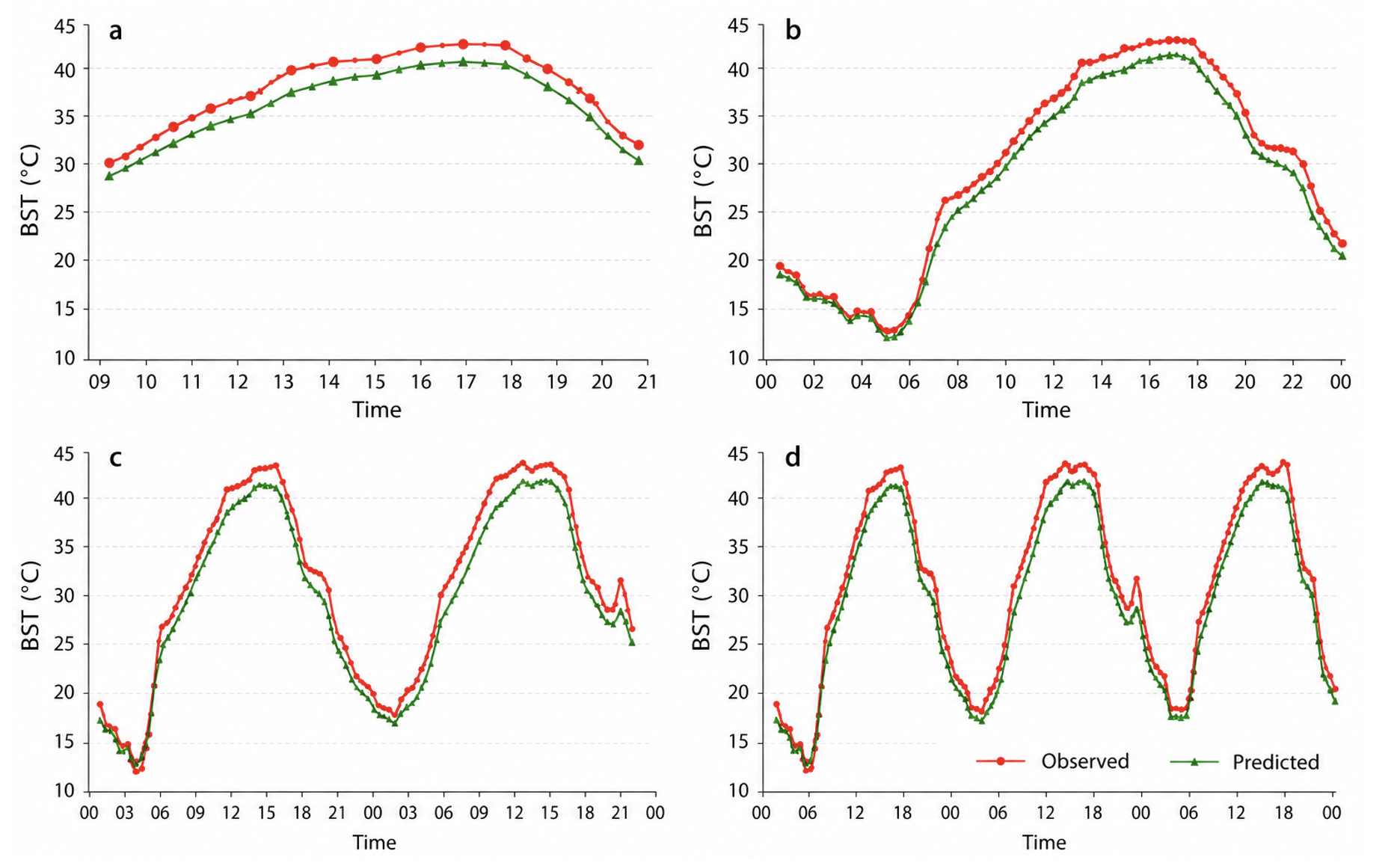}
    \caption{Berry temperature ($T_b$) forecasts with feed-forward attention mechanism and long-short term memory (FAM-LSTM) using open-field weather station dataset for a) 12-h, b) 24-h, c) 48-h, and d) 72-h.}
    \label{fig:pred-open}
\end{figure}

Figure \ref{fig:pred-open} illustrates the inference performance of the FAM-LSTM model for \Tb forecasting over the subsequent 72 h. The observed and predicted \Tb trajectories demonstrate that the model consistently captured the diurnal temperature dynamics across all forecast horizons. However, a systematic underestimation of peak daytime \Tb was observed when using the open-field weather dataset, particularly during periods of elevated berry temperatures. This underestimation may be attributed to differences between open-field meteorological conditions and the localized vineyard microclimate. Typically, a standard weather-station measurements may not fully capture canopy-level radiation and thermal conditions experienced by the berries.

\subsection{Performance of in-vineyard dataset for berry temperature forecasting}

In-vineyard weather-driven models consistently achieved superior prediction accuracy compared to their open-field counterparts across all models and forecast horizons, as summarized in Table \ref{tab:in_vineyard_performance}. The FAM-LSTM model trained on in-vineyard data achieved the best overall performance at all horizons. At the 15-min horizon, in-vineyard FAM-LSTM obtained MAE = 0.51 \degC and MAPE = 2.48\%, compared with MAE = 0.58 \degC and MAPE = 2.65\% for the open-field FAM-LSTM, which is an improvement of 12.1\% in MAE. At the 72-h horizon, in-vineyard FAM-LSTM achieved MAE = 1.55 \degC and MAPE = 7.68\%, compared with MAE = 1.70 \degC and MAPE = 8.09\% for the open-field dataset, representing an 8.8\% improvement in MAE. The consistent performance advantage of in-vineyard models across all modeling frameworks can be attributed to the closer physical coupling between canopy meteorological conditions and $T_b$. In-vineyard sensors captured the local thermal, radiative, and moisture environment that directly governs berry energy balance, whereas open-field stations measure conditions at a standardized reference height over bare ground, introducing systematic biases due to the differences in surface albedo and roughness length. However, the open-field FAM-LSTM model still achieved acceptable prediction accuracy across all horizons (MAPE < 9\% at 72 hours), suggesting that widely available regional meteorological data can serve as a reasonable surrogate for in-vineyard microclimate inputs when dedicated canopy-level sensors are not installed. This is a practically significant finding, as open-field data from a wine grape growing region can enable reliable \Tb forecasting for growers without in-vineyard monitoring infrastructure.

\begin{table}[htbp]
\centering
\caption{Prediction performance of evaluated models using in-vineyard weather inputs across multiple forecast horizons.}
\label{tab:in_vineyard_performance}

\setlength{\tabcolsep}{6pt}
\renewcommand{\arraystretch}{1.15}

\begin{tabular}{llccccccc}
\toprule
\textbf{Model} & \textbf{Metric}
& \textbf{$t+1$}
& \textbf{$t+4$}
& \textbf{$t+24$}
& \textbf{$t+48$}
& \textbf{$t+96$}
& \textbf{$t+192$}
& \textbf{$t+288$} \\

&
& \textbf{15 min}
& \textbf{1 h}
& \textbf{6 h}
& \textbf{12 h}
& \textbf{24 h}
& \textbf{48 h}
& \textbf{72 h} \\
\midrule

\multirow{3}{*}{RF}
& MAE  & 1.01 & 1.40 & 1.70 & 1.78 & 1.95 & 2.02 & 2.17 \\
& RMSE & 1.39 & 1.81 & 2.08 & 2.15 & 2.40 & 2.34 & 2.84 \\
& MAPE & 5.02 & 6.68 & 7.98 & 8.11 & 9.45 & 9.56 & 10.35 \\
\addlinespace

\multirow{3}{*}{RNN}
& MAE  & 0.85 & 1.10 & 1.54 & 1.68 & 1.72 & 1.69 & 1.71 \\
& RMSE & 1.09 & 1.37 & 1.85 & 2.01 & 2.10 & 2.15 & 2.15 \\
& MAPE & 4.18 & 5.23 & 7.54 & 7.82 & 7.78 & 8.90 & 8.69 \\
\addlinespace

\multirow{3}{*}{GRU}
& MAE  & 0.84 & 1.16 & 1.45 & 1.58 & 1.63 & 1.78 & 1.86 \\
& RMSE & 1.08 & 1.48 & 1.84 & 1.97 & 2.06 & 2.12 & 2.20 \\
& MAPE & 4.16 & 5.88 & 7.03 & 7.65 & 8.14 & 8.92 & 9.76 \\
\addlinespace

\multirow{3}{*}{LSTM}
& MAE  & 0.53 & 1.07 & 1.40 & 1.52 & 1.47 & 1.67 & 1.79 \\
& RMSE & 0.76 & 1.31 & 1.68 & 1.90 & 1.84 & 2.09 & 2.17 \\
& MAPE & 2.57 & 5.22 & 6.92 & 7.36 & 7.07 & 8.70 & 9.09 \\
\addlinespace

\multirow{3}{*}{\textbf{FAM-LSTM}}
& MAE
& \textbf{0.51}
& \textbf{0.98}
& \textbf{1.31}
& \textbf{1.39}
& \textbf{1.40}
& \textbf{1.59}
& \textbf{1.55} \\

& RMSE
& \textbf{0.71}
& \textbf{1.24}
& \textbf{1.60}
& \textbf{1.70}
& \textbf{1.72}
& \textbf{1.91}
& \textbf{1.87} \\

& MAPE
& \textbf{2.48}
& \textbf{4.91}
& \textbf{6.43}
& \textbf{7.27}
& \textbf{6.97}
& \textbf{7.88}
& \textbf{7.68} \\

\bottomrule
\end{tabular}

\vspace{2mm}

\footnotesize
\textit{Note:} MAE and RMSE are expressed in \degC; MAPE is expressed in \%.
Bold values indicate the best performance for each metric and forecast horizon.

\end{table}

\begin{figure}[!htbp]
    \includegraphics[width=\linewidth]{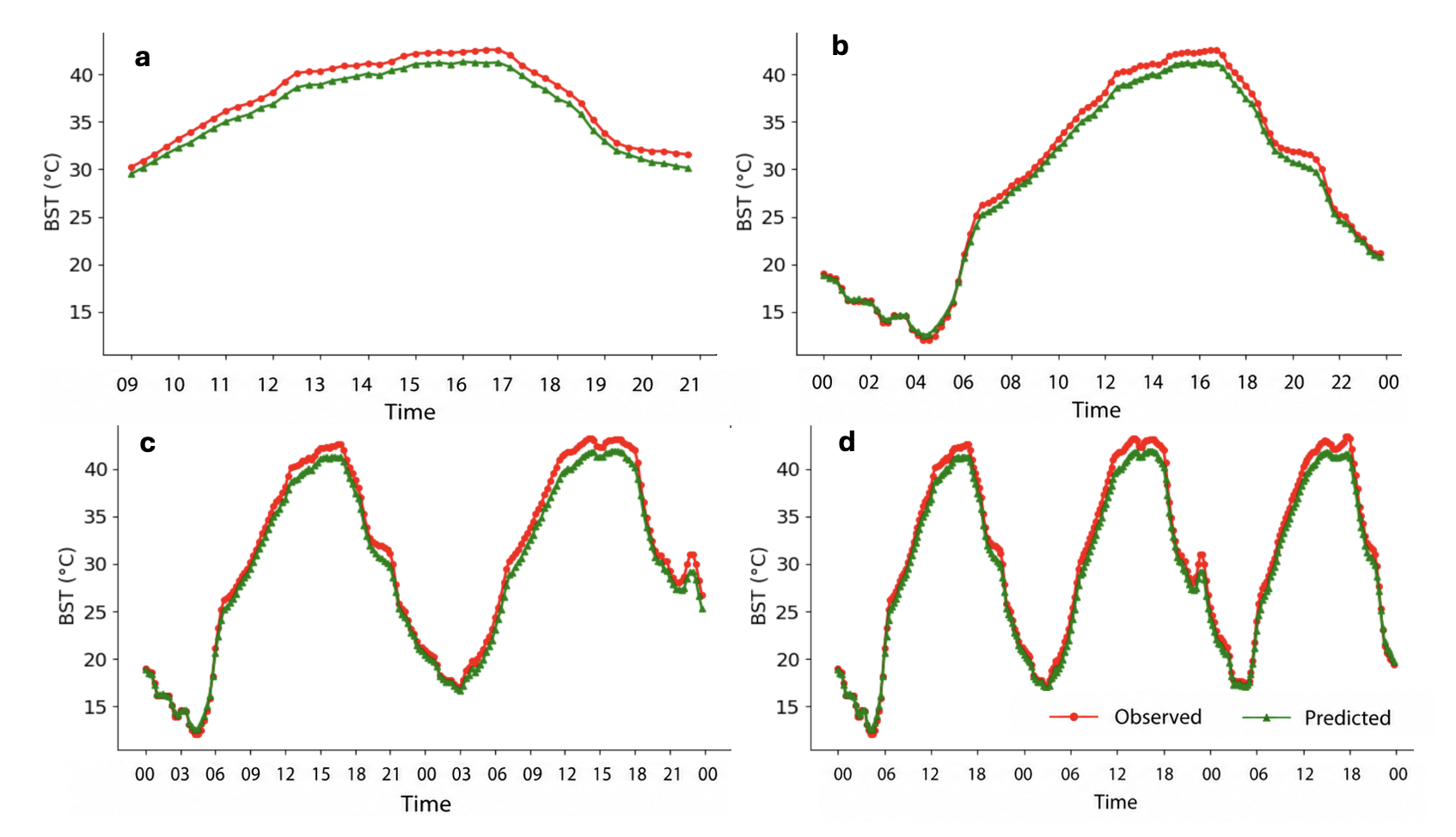}
    \caption{Berry temperature ($T_b$) forecasts with feed-forward attention mechanism and long-short term memory (FAM-LSTM) using in-vineyard dataset for a) 12-h, b) 24-h, c) 48-h, and d) 72-h.}
    \label{fig:in-vine}
\end{figure}

Figure \ref{fig:in-vine} illustrates the FAM-LSTM predictions using in-vineyard microclimate inputs over the 72-h forecast horizon. The predicted \Tb closely followed the observed diurnal patterns across all forecast horizons. Compared with predictions using the open-field weather dataset, the underestimation of peak \Tb was substantially reduced but not completely eliminated. This improved agreement highlights the advantage of localized in-vineyard measurements in capturing the microclimatic conditions experienced by grape berries, thereby improving the accuracy of multi-step \Tb forecasting.

\subsection{Temporal error analysis across diurnal and horizon zones}

To characterize how prediction error varied across diurnal thermal regimes and progressive forecast sub-periods, detailed zone-level error analyses were conducted for the FAM-LSTM model under both open-field and in-vineyard scenarios (Tables \ref{tab:diurnal_openfield} and \ref{tab:diurnal_invineyard}, respectively). Three zones were analyzed within each 24-hour forecast sub-period: daily maximum temperature (peak temperature), the daytime warming window (11:00-18:00), and the overnight and morning cooling window (18:00-11:00 the following day).

For the open-field FAM-LSTM, errors were consistently lowest during the overnight and morning cooling period (18:00-11:00) and highest during the daytime warming window (11:00-18:00). In the 0–24 h sub-period, the nighttime zone achieved MAE = 0.86 \degC and MAPE = 3.44\%, while the daytime window produced MAE = 1.66 \degC and MAPE = 4.12\%, and the peak temperature error was MAE = 1.66 \degC and MAPE = 3.85\%. The peak temperature prediction error increased from MAE = 1.66 \degC in the first day to MAE = 2.03 \degC in the 48–72 h period, representing a 22.4\% increase. The higher prediction errors during the daytime window can be attributed to the greater variability in vineyard microclimatic conditions, driven by rapidly changing solar radiation, wind, and turbulent heat exchange. These dynamic conditions can produce substantial differences between berry-level thermal conditions and measurements from an open-field weather station, making daytime \Tb more difficult to predict. In contrast, nighttime \Tb is primarily influenced by longwave radiative cooling and generally exhibits smoother temporal variations, resulting in lower prediction errors when using open-field weather-station inputs. Prediction errors also progressively increased across the three forecast periods (0-24 h, 24-48 h, and 48-72 h) for all evaluated zones (peak temperature, 11:00-18:00, and 18:00-11:00), reflecting the increasing uncertainty associated with predicting \Tb farther into the multi-day forecast horizon.

\par In-vineyard FAM-LSTM models achieved substantially lower zone-level errors across all periods. In the 0-24 h sub-period, the peak temperature error was reduced by 29.2\% with in-vineyard data (MAE = 1.17 $^\circ$C; MAPE = 2.71\%) in comparison with open-field weather data (MAE = 1.66 $^\circ$C; MAPE = 3.85\%). Similarly, in day and nighttime zones, the error was reduced by 37.3\% and 36.0\% in the 0-24 h period using the in-vineyard data. This improvement implies that while the model predicts the diurnal patterns of \Tb using an open-field weather station, it requires in-vineyard data for accurate prediction of \Tb for effective heat stress management in vineyards. 

The zone-level analysis demonstrated the potential of the \Tb forecasting model for operational heat-stress monitoring using readily available open-field weather-station data. Peak \Tb prediction errors remained within approximately 2 \degC at the 72-h forecast horizon for temperatures exceeding 30 \degC, which is comparable to the air-temperature forecast errors reported for the National Blend of Models (NBM) and Numerical Weather Prediction (NWP) models at similar forecast horizons \citep{gaudet2024verification-9fa, mass2024pacific-d5c}. This level of accuracy supports the practical application of the model for advance heat-stress alerts and \Tb monitoring. However, further improvements in peak-temperature prediction would be necessary for real-time precision heat-stress mitigation, where accurate identification of critical temperature thresholds is essential for timely activation of cooling systems.

\begin{table}[htbp]
\centering
\caption{Diurnal zone error analysis for the open-field FAM-LSTM model 
across 24-h forecast sub-periods. MAE and RMSE are expressed in $^\circ$C, 
and MAPE is expressed in \%.}
\label{tab:diurnal_openfield}

\renewcommand{\arraystretch}{1.2}
\setlength{\tabcolsep}{8pt}

\begin{tabular}{llccc}
\toprule
\textbf{Forecast period} & \textbf{Zone} & 
\textbf{MAE} & \textbf{RMSE} & \textbf{MAPE (\%)} \\
\midrule

\multirow{3}{*}{0--24 h}
& Peak Temp (max)       & 1.66 & 1.66 & 3.85 \\
& 11:00--18:00          & 1.66 & 1.68 & 4.12 \\
& 18:00--11:00 (next)   & 0.86 & 1.01 & 3.44 \\
\midrule

\multirow{3}{*}{24--48 h}
& Peak Temp (max)       & 1.67 & 1.67 & 3.93 \\
& 11:00--18:00          & 1.78 & 1.82 & 4.29 \\
& 18:00--11:00 (next)   & 1.17 & 1.30 & 4.17 \\
\midrule

\multirow{3}{*}{48--72 h}
& Peak Temp (max)       & 2.03 & 2.03 & 4.69 \\
& 11:00--18:00          & 2.00 & 2.04 & 4.77 \\
& 18:00--11:00 (next)   & 1.40 & 1.54 & 5.01 \\
\bottomrule

\end{tabular}
\end{table}

\begin{table}[htbp]
\centering
\caption{Diurnal zone error analysis for the in-vineyard FAM-LSTM model 
across 24-h forecast sub-periods. MAE and RMSE are expressed in $^\circ$C, 
and MAPE is expressed in \%.}
\label{tab:diurnal_invineyard}

\renewcommand{\arraystretch}{1.2}
\setlength{\tabcolsep}{8pt}

\begin{tabular}{llccc}
\toprule
\textbf{Forecast period} & \textbf{Zone} &
\textbf{MAE} & \textbf{RMSE} & \textbf{MAPE (\%)} \\
\midrule

\multirow{3}{*}{0--24 h}
& Peak Temp (max)       & 1.17 & 1.17 & 2.71 \\
& 11:00--18:00          & 1.04 & 1.07 & 2.69 \\
& 18:00--11:00 (next)   & 0.55 & 0.64 & 2.66 \\
\midrule

\multirow{3}{*}{24--48 h}
& Peak Temp (max)       & 1.35 & 1.35 & 3.16 \\
& 11:00--18:00          & 1.30 & 1.32 & 3.32 \\
& 18:00--11:00 (next)   & 0.79 & 0.86 & 3.54 \\
\midrule

\multirow{3}{*}{48--72 h}
& Peak Temp (max)       & 1.57 & 1.57 & 3.68 \\
& 11:00--18:00          & 1.73 & 1.76 & 4.38 \\
& 18:00--11:00 (next)   & 0.87 & 0.97 & 3.82 \\
\bottomrule

\end{tabular}
\end{table}

\subsection{Practical implications for vineyard heat stress management}

\par The FAM-LSTM model developed in this study provides a decision-support framework for precision heat-stress management in viticulture by generating \Tb forecasts at 15-min intervals for next 72 h. This extended forecast horizon provides vineyard managers with sufficient lead time to plan and implement different heat stress-mitigation strategies. For instance, overhead evaporative cooling may require 12-24 h of advance planning to confirm water availability from irrigation districts, while deployment of shade nets may require additional time for labor and field operations. Advance \Tb forecasts can also support adjustments to irrigation schedules, particularly when evaporative cooling contributes additional water to the soil profile. Therefore, the 72-h forecast horizon provides actionable information that can facilitate coordinated planning of water resources, labor, irrigation, and other heat-mitigation interventions before potentially damaging berry temperatures occur.

\par The open-field weather data driven FAM-LSTM model achieved acceptable forecasting performance, with a MAPE below 9\% at the 72-h horizon, demonstrating its potential for meso- to regional-scale \Tb forecasting across grape-growing regions. The use of existing weather-station networks also reduces the need for capital investment towards in-vineyard sensor networks, potentially improving the accessibility of $T_b$-based heat stress decision support. This capital reduction is key for small- and medium-sized vineyard operations with limited resources for canopy-level instrumentation. Nevertheless, incorporating in-vineyard microclimate measurements further improved forecasting performance, especially during periods of peak daytime temperatures. The approximately 29\% reduction in peak \Tb prediction error achieved using in-vineyard inputs could improve the reliability of heat-stress alerts during the most thermally critical periods, where accurate prediction of temperature extremes is particularly important for timely mitigation decisions.

\par The practical importance of forecasting \Tb arises from the substantial differences that can occur between berry and ambient air temperatures, particularly during periods of high solar exposure. Previous studies that assumed berry temperature to be equivalent to air temperature may therefore underestimate the actual thermal stress experienced by the fruit \citep{bindi1997simple, due1993modelling-abc, wermelinger1991demographic-abc}. Spatially explicit modeling with Helios demonstrated that berry temperature deviations from ambient conditions vary with vineyard geometry, trellis configuration, solar exposure, and berry heat storage, and that accurately representing these deviations is particularly important for temperature-sensitive processes such as changes in berry chemical composition and sunburn injury \citep{PoncedeLen2021A3M}. These findings are consistent with previous studies showing that elevated berry temperatures can alter sugar and organic acid metabolism and inhibit anthocyanin accumulation \citep{spayd2002separation-abc}, ultimately affecting berry and wine quality \citep{reshef2019grape-abc}. Therefore, forecasting \Tb rather than relying solely on air temperature would be a more direct representation of fruit-level thermal exposure. The proposed FAM-LSTM framework can complement physical-based approaches by learning the nonlinear relationship between readily available meteorological variables and \Tb and providing forecasts up to 72 h in advance. Such forecasts could support early identification of potentially damaging thermal conditions and provide actionable lead time for heat-stress mitigation, while future integration of vineyard geometry, berry growth stage and size as well as canopy-exposure could further improve prediction of spatial variability in berry thermal stress. 

\par Future work should focus on extending and validating the proposed modeling framework across additional grape cultivars, vineyard locations, and canopy architectures with varying levels of berry sun exposure. Evaluating both red and white cultivars would also help assess model robustness under differences in berry thermal response and heat sensitivity. Integration of short-range NWP forecasts as exogenous inputs to the FAM-LSTM architecture could enable operational berry temperature forecasting based on predicted rather than observed meteorological conditions and potentially extend the actionable forecast horizon beyond 72 h. Furthermore, transfer learning approaches could be investigated to determine whether models developed at one vineyard can be efficiently adapted to new locations using limited site-specific data, thereby reducing data requirements and facilitating scalable deployment of berry temperature forecasting systems across diverse grape-growing regions.

\section{Conclusions}

The following are the conclusions of this study:

\begin{enumerate}
    \item The FAM-LSTM model consistently outperformed all benchmark models (e.g., LSTM, GRU, RNN, RF) across all prediction horizons upto 72 h and two practical scenarios, i.e., open-field and in-vineyard data availability as model inputs. The integration of feed-forward attention with LSTM recurrent encoding provides a robust and generalizable architecture for multi-step grape berry temperature forecasting.

    \item In-vineyard microclimate data consistently yielded superior model accuracy compared to open-field meteorological inputs across all models and horizons. In-vineyard FAM-LSTM achieving MAE reductions of 8.8-29.2\% relative to the open-field FAM-LSTM. Open-field FAM-LSTM achieved acceptably low errors (MAPE < 9\% at 72 hours), confirming the viability of using regional network-based weather station data as a surrogate input for \Tb forecasting in the absence of localized in-vineyard monitoring.

    \item Diurnal error analysis revealed that prediction uncertainty is highest during the daytime warming window (11:00--18:00) and at daily peak $T_b$, and increases systematically with forecast horizon, increasing from MAE = 0.5--1.7 \degC at 0–24 h to MAE = 1.4--2.0 \degC at 48–72 h for peak temperature predictions. Despite this temporal degradation, forecast accuracy at 72 hours remained within operationally meaningful bounds for heat stress decision support systems.
\end{enumerate}

\section*{Acknowledgements}
This research was partly funded by the Washington State Wine Commission, USDA-NIFA projects Award No. 2021-67021-35344, and \texttt{\#}0745. The authors thank the WSU AgWeatherNet program for provision of open-field meteorological data, Precision Agriculture Lab team, and the staff of the WSU Irrigated Agriculture Research and Extension Center, Prosser, WA, for field support during the 2023–2025 growing seasons.

\clearpage

\bibliographystyle{elsarticle-harv} 
\bibliography{references}

\end{document}